\documentclass{article}
\usepackage[margin=1in]{geometry}
\usepackage{authblk}
\usepackage{url}
\usepackage{hyperref}
\usepackage{setspace}
\usepackage{amsmath}
\usepackage{pdfpages}
\usepackage{multirow}
\usepackage{tabularx}
\usepackage{tabularray}
\usepackage{booktabs}
\usepackage{listings}
\usepackage{adjustbox}

\title{Evaluating Retrieval-Augmented Generation vs. Long-Context Input for Clinical Reasoning over EHRs}

\author[1]{Skatje Myers}
\author[2]{Dmitriy Dligach}
\author[3,4]{Timothy A. Miller}
\author[1]{Samantha Barr}
\author[1]{James Landefeld}
\author[5]{Yanjun Gao}
\author[1]{Matthew M. Churpek}
\author[1]{Anoop Mayampurath}
\author[1]{Majid Afshar}
\affil[1]{University of Wisconsin-Madison}
\affil[2]{Loyola University Chicago}
\affil[3]{Boston Children’s Hospital}

\affil[4]{Harvard Medical School}
\affil[5]{University of Colorado-Anschutz} 

\begin{document}

\maketitle

\begin{abstract}\textbf{Objective:} To evaluate whether retrieval-augmented generation (RAG) can serve as an efficient alternative to long-context prompting for clinical reasoning over electronic health records (EHRs). \\ \textbf{Methods:} We defined three EHR-based tasks that are replicable across health systems and vary in reasoning complexity: 1) extracting imaging procedures (modality, date, and anatomic site), 2) generating timelines of therapeutic antibiotic use, and 3) identifying the key diagnoses for a hospitalization. Using real inpatient clinical notes from a US academic health system, we evaluated three large language models (GPT-5.4-mini, Mistral Medium 3, DeepSeek V3.1) with varying amounts of provided context, comparing targeted retrieval to using the most recent clinical notes. \\ \textbf{Results:} For \textit{Imaging Procedures}, RAG strongly outperformed recent-note inputs and exceeded long-context performance (by 0.17-9.83 F1 across all models) using fewer than 8K tokens. Similar benefits were observed for \textit{Antibiotic Timelines}, where \textless8K of retrieved tokens matched long-context recent-notes performance (between -3.26 to +3.24 Jaccard).  Error analysis revealed that missing information in the clinical notes—often due to inter-hospital transfers—limited performance to some extent. However, performance on the \textit{Diagnosis Generation} task remains largely static across methods and models. \\ \textbf{Discussion:} RAG demonstrated strong token efficiency across tasks, with the clearest and most consistent gains observed for imaging extraction and antibiotic timeline reconstruction. Diagnosis generation proved the most challenging task, suggesting ceiling effects imposed by documentation variability and evaluation constraints. \\ \textbf{Conclusion:} Our results suggest that RAG remains a competitive and efficient approach for clinical tasks over large amounts of EHR, even as newer models become capable of handling increasingly longer amounts of text.
\end{abstract}

\section{Introduction}

Electronic health records (EHRs) contain comprehensive documentation of patient care, including critical information for diagnosis and treatment planning. However, the volume of clinical notes has increased substantially in recent years, driven in part by copy-paste practices, templated documentation, and regulatory pressures—a phenomenon often referred to as ``note bloat''. For example, nearly 1 in 5 patients arrive at the emergency department with a chart the size of Moby Dick (more than 200K words) \cite{patterson2024call}. As a result, clinicians must navigate increasingly lengthy and redundant records to locate key information.

Large language models (LLMs) can potentially alleviate this burden by assisting clinicians in quickly extracting information and reasoning over EHRs, and have demonstrated promising capabilities in clinical summarization \cite{van2024adapted} and question answering \cite{singhal2025toward}. However, the sheer volume of clinical documentation can exceed most LLMs' context window size. A practical approach is to provide the most recent notes, which may suffice for some tasks but risks omitting crucial information buried in earlier documentation.

\begin{figure*}[h]
\centering
\noindent\includegraphics[width=\textwidth]{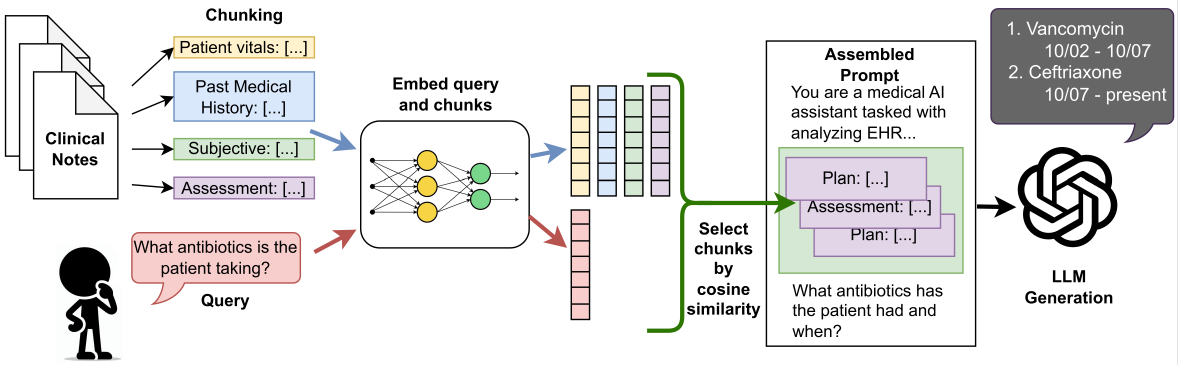}
\caption{Retrieval-augmented generation pipeline for clinical question answering over EHR.}
\label{fig:rag}
\end{figure*}

Retrieval-augmented generation (RAG) has emerged as a promising solution to using LLMs on long documents by retrieving only the most relevant text passages for a given task. Rather than processing entire patient charts, RAG systems can selectively extract pertinent clinical information to answer specific questions (Figure \ref{fig:rag}). This approach can potentially reduce computational costs, improve accuracy by eliminating noise, and mitigate the ``lost-in-the-middle'' effect \cite{liu-etal-2024-lost}, where model performance degrades when relevant information is buried within lengthy contexts. 

However, there has been limited empirical evaluation on the accuracy and token efficiency of this retrieval approach for tasks that require longitudinal reasoning over real-world EHR data. One barrier is the scarcity of large, annotated clinical datasets due to legal and ethical constraints regarding patient privacy. While the MIMIC datasets \cite{johnson2016mimic} have been further annotated for benchmarking a variety of natural language processing tasks, including question-answering, these data are restricted to the patients' ICU stay, as opposed to the full hospital course, limiting their potential for testing realistic use cases that stretch the token limitations of LLMs' processing abilities.

To address these gaps, we define three tasks that reflect different clinical reasoning demands and can be replicated in other health systems without labor-intensive manual annotation:
\begin{itemize}
    \item \textit{Imaging Procedures}: Produce a list of imaging procedures (including modality, date, and anatomical location) that occurred during a hospitalization from the raw clinical notes.
    \item \textit{Antibiotic Timelines}: Generate the timelines of therapeutic antibiotic use for patients with a life-threatening infection and, in many cases, have sepsis.
    \item \textit{Diagnosis generation}: Identify the key diagnoses relevant to the hospitalization.
\end{itemize}

The \textit{Imaging Procedures} task is a straightforward extractive task that requires the model to identify the imaging procedures performed on different days during the hospital stay. The \textit{Antibiotic Timelines} task requires not only identifying the antibiotics the patient was on and when they were discontinued, but also incorporating implicit medical reasoning to determine the indications for those antibiotics (e.g., prophylaxis, suppression, prior treatment, current treatment, etc.). The final task, \textit{Diagnosis Generation}, requires the most medical reasoning—the model is asked not just to list the diagnoses mentioned, but to determine which required active management and affected the care plan.

These tasks allow us to investigate the following questions: (a) given a limited token budget, does a simple, generalizable RAG approach improve efficiency and performance compared to providing an LLM with only the most recent notes?, and (b) to what extent can this baseline approach approximate the performance achieved by utilizing the full long-context window of modern models?

Using EHR data from the University of Wisconsin (UW) hospital, we evaluate three LLMs on these tasks using varying amounts of clinical context, including up to the models' full context window of 120K tokens.

Our findings suggest that while a simple RAG can provide substantial efficiency improvements over comparable amounts of recent clinical note tokens, this effect may be task-dependent. However, across all three tasks, we found that RAG achieved near-parity with the full context window using a fraction of the tokens, indicating that retrieval remains a competitive approach even as newer model architectures continue to extend context windows.

\section{Related Work}

Many existing question-answering datasets for EHRs focus on fact extraction. Datasets such as EmrQA \cite{pampari-etal-2018-emrqa} and DrugEHRQA \cite{bardhan-etal-2022-drugehrqa} are semi-automatically constructed by leveraging previous annotations from National NLP Clinical Challenges (n2c2) to transform them into question-answer pairs. For this type of data, template questions are constructed where the annotation can fill a slot, such as ``What is the dosage of $\vert$medication$\vert$?''. 

The MedAlign dataset \cite{fleming2024medalign} comprises clinician-generated instruction-answer pairs and longitudinal EHR data. Many of the instructions are yes/no questions that can be answered by retrieving a single piece of evidence (e.g. ``Does she smoke?''), but some instructions require the model to synthesize information across the EHR (e.g. ``Summarize this patient's cardiac history.''). However, evaluation on open-ended responses poses an ongoing challenge in NLP, with popular automatic metrics such as BLEU and ROUGE showing poor correlation with human judgment on natural language generation tasks in healthcare \cite{croxfordDevelopmentHumanEvaluation2024a}.

Retrieval-augmented generation has been used for a variety of tasks within the medical domain, including answering open-ended medical questions by retrieving from medical guidelines and journal articles \cite{zakka2024almanac} and assessing surgical fitness by retrieving from perioperative guidelines \cite{ke2025retrieval}. Alkhalaf et al. \cite{alkhalaf2024applying} used RAG to generate structured summaries by retrieving from EHR, querying for relevant text using the names of the summary fields (such as ``age'' and ``weight''). The MIRAGE benchmark \cite{xiong-etal-2024-benchmarking} is a collection of medical examination and biomedical research QA datasets to evaluate RAG systems. 

Outside the clinical domain, several studies have directly compared RAG against long-context prompting. Li et al. \cite{li-etal-2024-retrieval} found that  long-context models generally outperformed retrieval on general-domain question-answering benchmarks, though at substantially higher computational cost, motivating a hybrid approach that routes queries between the two methods. In contrast, Yu et al. \cite{yu2024defense} demonstrated that an order-preserving retrieval mechanism could exceed full long-context performance with far fewer input tokens, and the LaRA benchmark \cite{pmlr-v267-li25dv} concluded that the optimal choice depends on the model, context length, and task type. Notably, these comparisons rely on general-domain corpora such as novels and academic papers, with conclusions that do not straightforwardly transfer to EHRs and reasoning tasks grounded in clinical workflows rather than single-answer question answering.

Our work aims to fill these gaps in the prior literature by extending the RAG-versus-long-context comparison to real-world longitudinal EHR data, using reproducible tasks that require reasoning over a patient's hospital course.

\section{Methods}
\subsection{Data and Models}
We constructed datasets of 200 inpatient hospitalizations for each of our three tasks using data from the University of Wisconsin hospital system, comprising clinical notes from admission to discharge (daily progress notes, specialist consultations, imaging reports, etc.). Table \ref{tab:data} provides summary statistics. All hospitalizations were at least 7 days long and comprised at least 15,000 tokens of clinical notes. For the \textit{Imaging Procedures} and \textit{Diagnosis Generation} tasks, only the clinical notes \textit{prior} to the discharge summary are used to provide information to the LLM, to avoid leaking information from the hospital course or diagnosis sections of the discharge summary. For the \textit{Antibiotic Timelines} task, all included hospitalizations involved a consultation with Infectious Diseases, and only the notes prior to the consultation note are included in the data presented to the LLM.

The only structured EHR data provided to the system were the notes' timestamps and types (e.g., progress note, handoff, etc.).

\begin{table}[]
\centering
\begin{tblr}{
  row{5} = {c},
  row{6} = {c},
  cell{1}{2} = {c},
  cell{1}{3} = {c},
  cell{1}{4} = {c},
  cell{2}{2} = {c},
  cell{2}{3} = {c},
  cell{2}{4} = {c},
  cell{3}{2} = {c},
  cell{3}{3} = {c},
  cell{3}{4} = {c},
  cell{4}{2} = {c},
  cell{4}{3} = {c},
  cell{4}{4} = {c},
  cell{5}{1} = {r},
  cell{6}{1} = {r},
  hline{1,7} = {-}{0.08em},
  hline{2} = {1-4}{},
}
                         & Imaging  & Antibiotics & Diagnosis \\
Hospitalizations               & 200      & 200         & 200       \\
Mean notes per hospitalization & 110      & 145         & 111       \\
Tokens per hospitalization:    &          &             &           \\
mean                     & 74k      & 108k        & 75k       \\
range                    & 17k-401k & 16k-1.4m    & 20k-389k  
\end{tblr}
\caption{Dataset statistics for each task.}
\label{tab:data}
\end{table}

We evaluated three state-of-the-art LLMs capable of processing up to 128K tokens:
\begin{itemize}
    \item \textbf{GPT-5.4-mini} \cite{openaiIntroducingGPT54}
    \item \textbf{Mistral Medium 3} \cite{mistralMediumLarge}
    \item \textbf{DeepSeek V3.1} \cite{deepseekDeepSeekV31Release}
\end{itemize}

\subsection{RAG System}

For each patient hospitalization, clinical notes were segmented into overlapping 128-token chunks, with a sliding window of 20. We embedded these chunks using two models: BGE-en-large-v1.5 \cite{bge_embedding}, a general-purpose BERT-based embedding model trained through contrastive learning, and Qwen3-Embedding-8B \cite{qwen3embedding}. We included BGE-en-large-v1.5 based on findings from Myers et al. \cite{myers2025lessons}. This ablation study of embedding models and pooling strategies for EHR retrieval found that BGE-en-large-v1.5 significantly outperformed general-domain and biomedical-domain alternatives across several related retrieval tasks over EHR. We additionally evaluated Qwen-3-Embedding to assess whether a more recent, larger-scale embedding model offers further gains for these tasks.

For each task, we manually crafted a simple query for retrieving relevant passages (e.g. ``What are the patient's diagnoses?''). Each query was prepended with the model's recommended instruction prefix: ``Represent this sentence for searching relevant passages:'' for BGE-en-large-v1.5, and ``Instruct: Given a web search query, retrieve relevant passages that answer the query\textbackslash nQuery: '' for Qwen3-Embedding-8B. We used cosine similarity between the query and each chunk to retrieve the top-N most relevant passages (N = 20, 40, 60). These chunks were inserted into the instruction prompt (provided in Appendix B) and passed to the LLM.

We compared this retrieval configuration to a baseline approach of providing the most recent clinical notes in comparable amounts (2560, 5120, or 7680 tokens of EHR text) and long-context inputs with up to 64K or 120K EHR tokens. References to these token amounts throughout this study should be understood as an \textit{upper bound}, as some encounters consist of fewer tokens, reflecting the underlying hospitalization distribution. Token counts reported throughout this study were computed using an independent tokenizer rather than each model's native tokenizer. For each token amount evaluated (e.g., 20 chunks of 128 tokens, 64K tokens), this ensured that all models received identical input text, since tokenizers can produce different counts for the same content.

Performance on the tasks was evaluated using either the F1 or Jaccard index, as described in the following sections, and we assessed comparative performance between the RAG and non-RAG approaches across increasing numbers of tokens by calculating the area under the curves and reporting the normalized area difference. Specifically, for each method we plotted task performance as a function of input token count and interpolated between the measured points to obtain continuous curves over the range of token budgets shared by both methods. We then computed the area under each curve using the trapezoidal rule and defined the normalized area difference as:
\begin{equation}
\text{Normalized Area Difference} = \frac{\text{AUC}_{\text{RAG}} - \text{AUC}_{\text{Recent Notes}}}
     {\text{AUC}_{\text{Recent Notes}}} \times 100\%
\end{equation}
A positive value indicates that RAG outperformed the recent-notes baseline across the shared token range, with the magnitude reflecting the relative size of the gap. A value of 100\% indicates that the area under the RAG curve was twice that of the Recent Notes curve over the evaluated range, while a negative value indicates that the recent-notes baseline performed better on average.

\subsection{Task 1: Imaging Procedures}

The \textit{Imaging Procedures} task involves extracting structured information from unstructured clinical notes about diagnostic imaging procedures. We focused on five common imaging modalities: Magnetic Resonance Imaging (MRI), Computed Tomography (CT), Ultrasound, X-ray, and Nuclear Medicine (NM) Imaging. The model was prompted to produce a list of imaging procedures performed during the hospitalization, specifying the modality, anatomical location, and date. Relevant passages were retrieved using a query string listing relevant procedures: ``X-ray, CT, MRI, ultrasound, NM imaging, echocardiogram, fluoroscopy''.

As a gold standard for this task, we used tabular procedure records from the EHR. We mapped these procedure descriptions to imaging modality and anatomical site using simple rules and regular expressions, and manually verified the 934 resultant mappings. Our test set contains 66 unique combinations of modality and location. For example:

\begin{verbatim}
   X-RAY CHEST 2 VIEWS
        modality: "X-ray"
        location: "chest"
   CT LUMBAR SPINE W/O IV CONTRAST
        modality: "CT"
        location: "lumbar spine"
\end{verbatim}

Evaluation metrics are reported for three levels of strictness:
\begin{itemize}
    \item \textsc{Modality+Date+Location}
    \item \textsc{Modality+Date}
    \item \textsc{Modality+Date(±1 day)}
\end{itemize}

In the lattermost case, we allow for a reasonable tolerance in the predicted date due to observed variation in the reported metadata times for the procedure and note. For example, the timestamp for the note may reflect the date it was filed into the system, rather than the date it was actually written.

It should also be noted that the anatomical location is not normalized, other than for capitalization. For example, predicting ``spine'' for the above example would not be deemed a positive match.

\subsection{Task 2: Antibiotic Timelines}

This task emulates work performed by Infectious Diseases (ID) physicians, documenting the antibiotic regimen for an active infection in their consultation note. When these specialists are consulted, they document the history of the present illness, including lab results and medications, and outline a treatment plan. This note typically contains a ``History of Anti-Infectives'' section, where they list the antibiotics that have been used to treat the infection of concern, omitting prophylactic or non-relevant anti-infectives. For example:
\begin{verbatim}
       Vancomycin: 1/16-present
       Ceftriaxone: 1/17-present
\end{verbatim}

These medication names and date ranges were manually annotated by ID specialists after reviewing the patient's chart and served as our ground truth for this task, after extraction using regular expressions. No notes authored by ID physicians were included in the data that was presented to the model, and only notes that were written prior to the ground truth note were made available. Relevant passages were retrieved using the query ``What antibiotics has the patient taken?''.

We evaluated system accuracy with the following metrics:
\begin{itemize}
    \item \textsc{Medications (name only)}: Classification accuracy of only the medications, disregarding timespans.
    \item \textsc{Timespan Overlap}: The overlap between the predicted and gold date ranges for each antibiotic, reported using the Jaccard index. A value of 1 indicates an exact match; 0 indicates no overlap, missing a medication entirely, or including a medication not present in the gold standard. Values are averaged over the dataset.
\end{itemize}

The medications in both the generated predictions and the gold data are normalized to their ingredients using the RxNorm \cite{nelson2011normalized} API provided by the National Library of Medicine and a handful of manual rules for edge cases, such as typos. This approach allows for accurate matching of generic and brand-name medications, such as Zosyn (piperacillin and tazobactam).

As a baseline, we used a rule-based approach to directly extract the time ranges for all medications of the ``anti-infective'' therapeutic class from the list of administered medications using the EHR's medication administration record (MAR), a tabular form ubiquitous in EHRs for tracking all medications and infusions. However, this list of medications includes those used to treat other conditions that were not the focus of the ID consultation. By formulating this task to replicate the ID specialists' work, rather than on replicating structured data as the \textit{Imaging Procedures} did, this task requires an additional level of medical reasoning to accurately conform to inclusion criteria.

\subsection{Task 3: Diagnosis Generation}
The goal of this task is to generate a list of diagnoses for a given hospitalization that is of primary relevance to the clinician.

We drew from two EHR sources to construct our gold labels for each hospitalization: 
\begin{itemize}
    \item \textsc{Discharge Summary}: The free text from the discharge summary that lists the primary and secondary diagnoses.
    \item \textsc{Billing Codes}: The lists of International Classification of Diseases (ICD-10) codes from the structured EHR for the hospitalization, manually annotated by billing coders.
\end{itemize}

The diagnoses annotated by the coders are guaranteed to be documented within the notes, provide a high degree of specificity, and are normalized to a standard vocabulary, but these lengthy lists often include diagnoses that are not necessarily considered to be of key importance to clinicians, such as a history of smoking, obesity, or minor issues such as a contusion. On the other hand, the discharge summary may not list specific diagnoses, such as noting post-surgical status (e.g., ``S/p kidney transplant'') or leaving some diagnoses undocumented because they can be inferred. For example, documenting that the patient is post-kidney transplant and has complications, but not that they have kidney disease.

While discharge summaries reflect the information of most clinical relevance to the clinicians providing treatment, the billing codes lend themselves better to validation due to their standardization. To produce a more balanced representation, we instructed Gemma-3-32B \cite{blogIntroducingGemma} to filter the billing ICD code lists to only the entries that reflect the clinician's primary foci for the hospital stay based on the text from the discharge summary. This \textsc{Filtered} list of ICD-10 codes serves as our primary evaluation target and the instruction prompt was designed to elicit this list by outlining inclusion and exclusion criteria for diagnoses (e.g., include acute conditions requiring ICU care, exclude stable chronic conditions or irrelevant historical diagnoses). Manual validation of a subset of these filtered lists was performed, as detailed in Appendix D, finding that they largely correlate with clinician judgment.

To enable classification evaluation, the free text generated by the LLM and from the discharge summary need to be normalized to ICD-10 codes. For this process, we trained the state-of-the-art SNOBERT \cite{snobert} architecture to extract Systematized Nomenclature of Medicine (SNOMED) concepts from the text\footnote{Training details are provided in Appendix C} and used the mappings provided by the SNOMED Clinical Terms data release to convert them to ICD-10. A manual assessment of the accuracy of this process is described in Appendix E.

ICD-10 is a hierarchical vocabulary, ranging from broad concepts (e.g. ``Anemia, unspecified'' [D64.9]) to highly specific (e.g. ``Age-related osteoporosis with current pathological fracture, right shoulder, initial encounter for fracture'' [M80.011A]). Due to this high granularity, evaluating this task requires a more fuzzy matching technique, rather than evaluating classification accuracy on the ICD codes themselves. For this purpose, we employed the Healthcare Cost and Utilization Project's Clinical Classifications Software Refined (CCSR) \cite{ahrqClinicalClassifications}. The CCSR provides a mapping from ICD-10 codes to about 530 clinically relevant categories. 

CCSR is a many-to-many mapping, which enables mapping very fine-grained ICD codes such as ``Hypertensive chronic kidney disease'' to multiple CCSR categories: ``Hypertension with complications and secondary hypertension" and ``Chronic kidney disease''. This approach allowed us to consider predicted diagnoses to be a match even if the LLM split them into ``Hypertension'' and ``Chronic kidney disease''.

Some broader non-billable ICD codes are not included in the CCSR mapping (e.g. ``Hypotension'' [I95]). In these cases, we used the set intersection of the CCSR categories that the ICD code's \textit{subcategories} (e.g., I95.3, I95.89, etc.) are mapped to.

\section{Results}
\subsection{Task 1: Imaging Procedures}

\begin{figure*}[h]
\centering
\noindent\includegraphics[width=\textwidth]{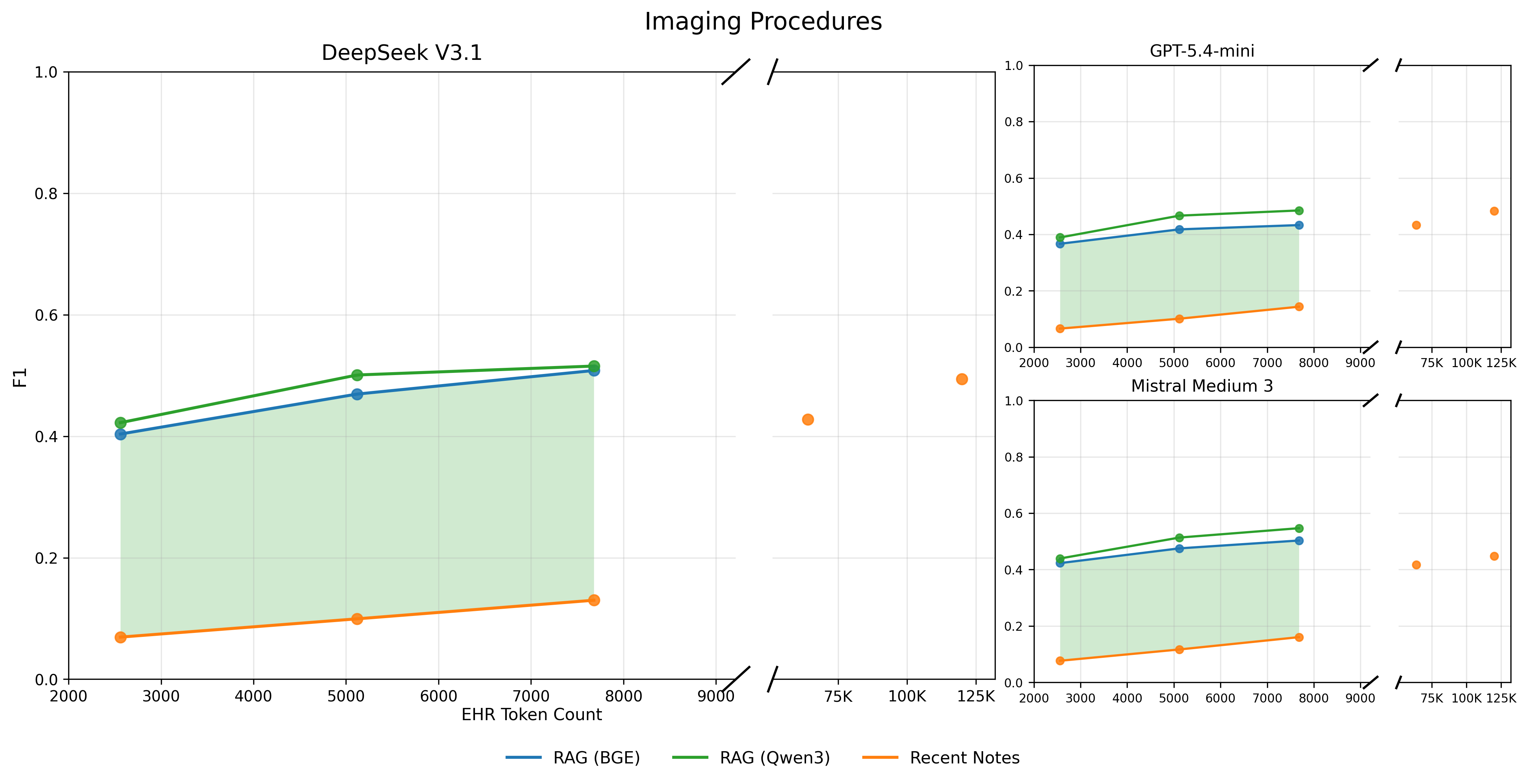}
\caption{\textit{Imaging Procedures}: F1 scores for the three models using the \textsc{Modality+Date+Location} evaluation method, across varying amounts of provided EHR tokens. The area between the curves used to calculate normalized area difference for RAG (BGE) is shown highlighted in green.}
\label{fig:imaging}
\end{figure*}

Across all three models and evaluation methods, RAG yielded dramatic performance improvements compared to comparable token amounts of recent notes for generating a list of the imaging procedures over the hospitalization. In Figure \ref{fig:imaging}, we show the classification performance for \textsc{Modality+Date+Location} across varying amounts of provided EHR context. We calculated the normalized area difference between the curves for the overlapping token amounts, presented in Table \ref{tab:imaging_results}. We found at minimum a 2.5-fold performance gain against using similar amounts of the most recent notes. These results also demonstrate that targeted retrieval of passages can closely approach the performance of utilizing the full context window with only a fraction of the tokens: Using only 60 chunks retrieved with Qwen3-Embedding, GPT-5.4-mini, Mistral Medium 3, and DeepSeek V3.1 exceeded the performance of using 120K tokens of recent notes by 0.17, 9.83, and 2.15, respectively. 

\begin{table}[h]
\centering

\resizebox{0.8\linewidth}{!}{%
\begin{tblr}{
  cell{1}{2} = {c=2}{c},
  cell{1}{4} = {c=2}{c},
  cell{1}{6} = {c=2}{c},
  cell{3}{2} = {c},
  cell{3}{4} = {c},
  cell{3}{6} = {c},
  cell{4}{2} = {c},
  cell{4}{4} = {c},
  cell{4}{6} = {c},
  cell{5}{2} = {c},
  cell{5}{4} = {c},
  cell{5}{6} = {c},
  hline{1,6} = {-}{0.08em},
  hline{2} = {2-7}{},
  hline{3-5} = {-}{},
}
                                                 & GPT-5.4-mini &          & Mistral Medium 3 &          & DeepSeek V3.1 &          \\
                                                 & BGE          & Qwen3    & BGE              & Qwen3    & BGE           & Qwen3    \\
\textsc{Modality+Date+Location} & 296.80\%     & 338.40\% & 298.03\%         & 327.02\% & 364.86\%      & 387.16\% \\
\textsc{Modality+Date}          & 263.27\%     & 281.38\% & 266.36\%         & 279.80\% & 311.08\%      & 321.15\% \\
\textsc{Modality+Date(±1 day)}  & 265.44\%     & 273.76\% & 251.52\%         & 256.40\% & 297.92\%      & 294.63\% 
\end{tblr}
}
\caption{Normalized area difference between the RAG and Recent Notes curves for the \textit{Imaging Procedures} task.}
\label{tab:imaging_results}
\end{table}

A complete listing of precision, recall, and F1 scores can be found in Appendix A.

\subsection{Task 2: Antibiotic Timelines}

\begin{figure*}[h]
\centering
\noindent\includegraphics[width=\textwidth]{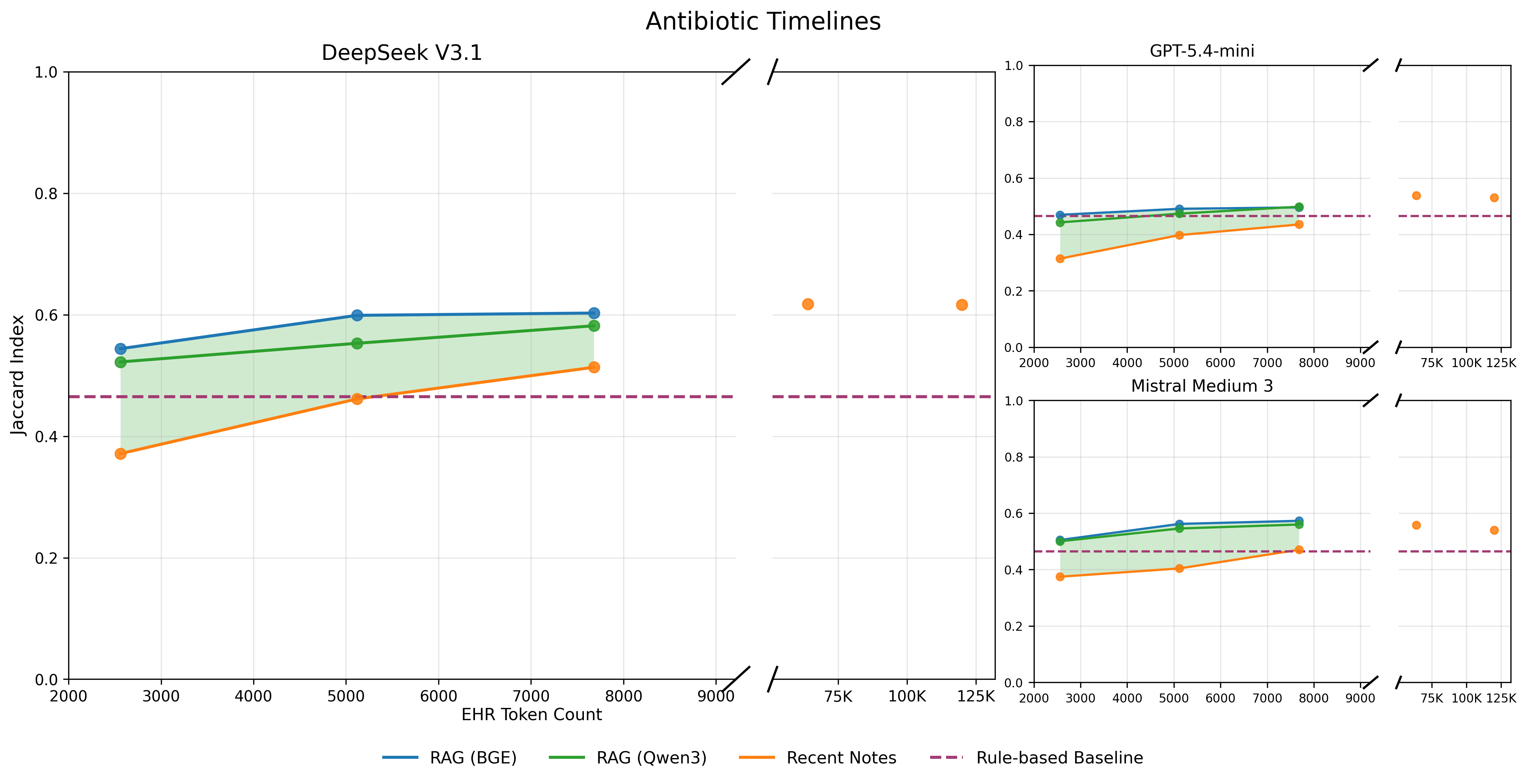}
\caption{\textit{Antibiotic Timelines}: Average Jaccard index for the three models using the \textsc{Timespan Overlap} evaluation method, across varying amounts of provided EHR tokens. The area between the curves used to calculate normalized area difference for RAG (BGE) is shown highlighted in green.}
\label{fig:antibiotics}
\end{figure*}

Figure \ref{fig:antibiotics} shows the performance of the models for \textsc{Timespan Overlap}. The RAG approach consistently exceeds the rule-based baseline and demonstrates close performance to using 120K tokens of recent notes (using 60 chunks retrieved with BGE, GPT-5.4-mini: -3.51, Mistral Medium 3: +3.24, DeepSeek V3.1: -1.34). The performance of the RAG approach shows only slight gains with increased retrieved text. The models show a consistent 22.23\%-33.16\% improvement in the normalized area difference between retrieval and recent-note methods for generating antibiotic timespans, as shown in Table \ref{tab:antibiotics}.

For all models, the average Jaccard index drops slightly when increasing the provided context from 64K to 120K tokens (GPT-5.4-mini: -0.78, Mistral Medium 3: -1.68, DeepSeek V3.1: -0.16).

On the task of predicting \textsc{Medications (name only)}, the RAG approach slightly outperforms using 120K tokens of recent notes with only 60 chunks retrieved by BGE (GPT-5.4-mini: +1.61, Mistral Medium 3: +5.56, DeepSeek V3.1: +0.34. A complete listing of Jaccard index, precision, recall, and F1 scores can be found in Appendix A.

Error analysis for this task draws attention to one of the limitations to be encountered when designing tasks for longitudinal EHR. In examining 70 encounters, we found that in 22.4\% of the gold medications, the information needed to generate the gold standard medication and precise timespan is not present in the full clinical notes. Most often, this occurs when the patient was transferred from another health system. This incomplete picture of a patient's history is hard to avoid when constructing datasets to capture longitudinal EHR, as patients don't exclusively visit a single healthcare system and healthcare data governance creates barriers to accessing this external information. The simple retrieval method we used also leaves further room for improvement—when retrieving 20 chunks using BGE, relevant information that could have improved performance was missed for 32\% of gold medications.

\begin{table}[h]
\centering
\resizebox{0.8\linewidth}{!}{%
\begin{tblr}{
  cell{1}{2} = {c=2}{c},
  cell{1}{4} = {c=2}{c},
  cell{1}{6} = {c=2}{c},
  cell{3}{2} = {c},
  cell{3}{4} = {c},
  cell{3}{6} = {c},
  cell{4}{2} = {c},
  cell{4}{4} = {c},
  cell{4}{6} = {c},
  hline{1,5} = {-}{0.08em},
  hline{2-3} = {2-7}{},
  hline{4} = {-}{},
}
                                                  & GPT-5.4-mini &         & Mistral Medium 3 &         & DeepSeek V3.1 &         \\
                                                  & BGE          & Qwen3   & BGE              & Qwen3   & BGE           & Qwen3   \\
\textsc{Medications (name only)} & 21.30\%      & 17.45\% & 30.00\%          & 26.75\% & 27.81\%       & 24.84\% \\
\textsc{Timespan Overlap}        & 26.04\%      & 22.23\% & 33.16\%          & 30.12\% & 29.72\%       & 22.27\% 
\end{tblr}
}
\caption{Normalized area difference between the RAG and Recent Notes curves for the \textit{Antibiotic Timelines} task.}
\label{tab:antibiotics}
\end{table}

\subsection{Task 3: Diagnosis Generation}
\begin{figure*}[h]
\centering
\noindent\includegraphics[width=\textwidth]{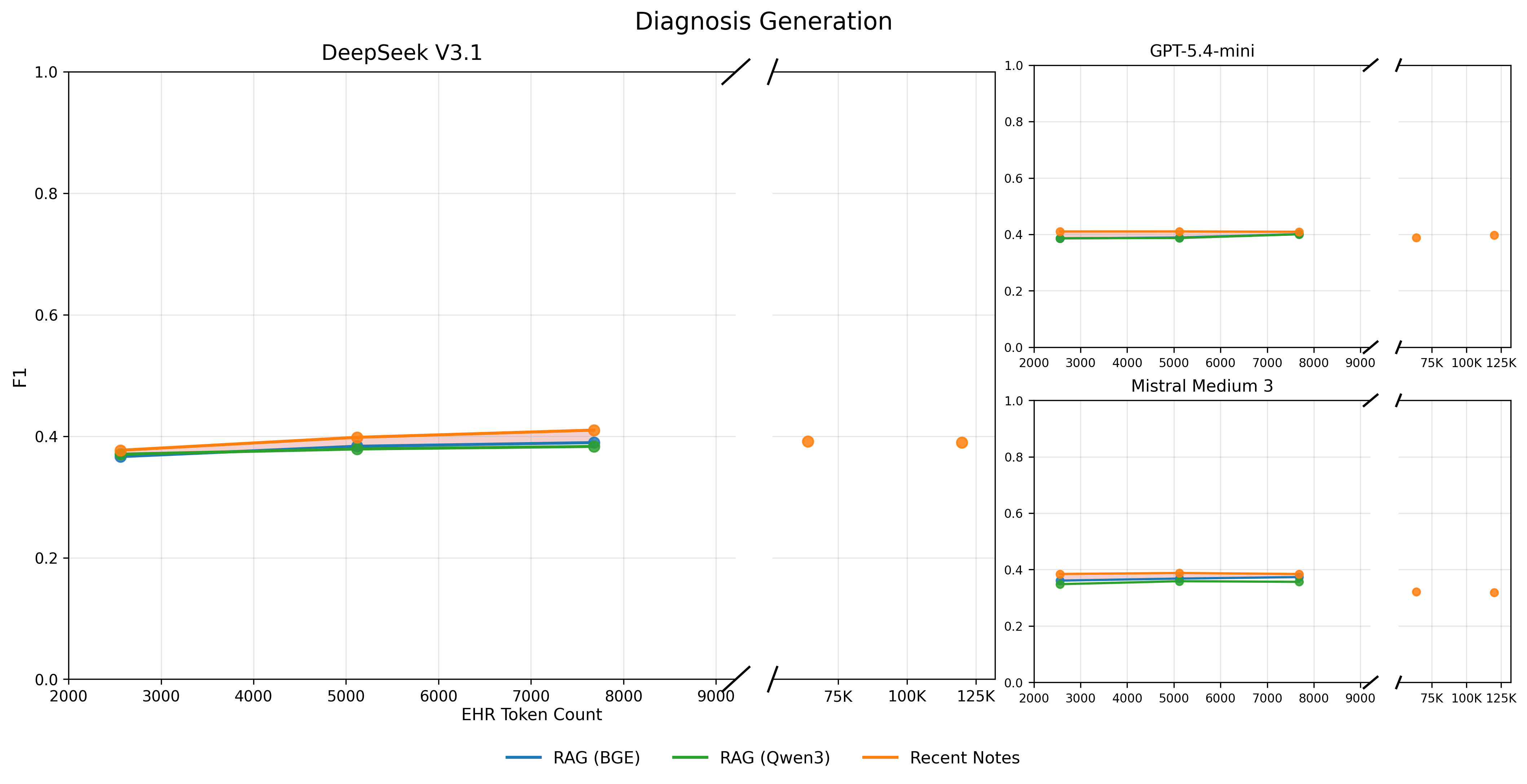}
\caption{\textit{Diagnosis Generation}: F1 scores for the three models using the \textsc{Filtered} evaluation method, across varying amounts of provided EHR tokens. The area between the curves used to calculate normalized area difference for RAG (BGE) is shown highlighted in red.}
\label{fig:diagnosis}
\end{figure*}

\begin{table}[h]
\centering
\resizebox{0.8\linewidth}{!}{%
\begin{tblr}{
  cell{1}{2} = {c=2}{c},
  cell{1}{4} = {c=2}{c},
  cell{1}{6} = {c=2}{c},
  cell{3}{2} = {c},
  cell{3}{4} = {c},
  cell{3}{6} = {c},
  cell{4}{2} = {c},
  cell{4}{4} = {c},
  cell{4}{6} = {c},
  cell{5}{2} = {c},
  cell{5}{4} = {c},
  cell{5}{6} = {c},
  hline{1,6} = {-}{0.08em},
  hline{2-3} = {2-7}{},
  hline{4-5} = {-}{},
}
                                            & GPT-5.4-mini &         & Mistral Medium 3 &          & DeepSeek V3.1 &         \\
                                            & BGE          & Qwen3   & BGE              & Qwen3    & BGE           & Qwen3   \\
\textsc{Billing codes}     & 2.90\%       & 7.62\%  & 7.23\%           & 8.52\%   & 4.33\%        & 7.81\%  \\
\textsc{Discharge summary} & -6.71\%      & -5.34\% & -10.32\%         & -11.50\% & -7.02\%       & -6.88\% \\
\textsc{Filtered}          & -4.67\%      & -4.86\% & -4.80\%          & -7.90\%  & -3.82\%       & -4.53\% 
\end{tblr}
}
\caption{Normalized area difference between the RAG and Recent Notes curves for the \textit{Diagnosis Generation} task.}
\label{tab:diag}
\end{table}

Unlike the previous two tasks, we do not see a consistent improvement in performance for using RAG compared to comparable amounts of recent notes, shown in Figure \ref{fig:diagnosis} and Table \ref{tab:diag}, but actually a slight decrease, other than evaluating against the \textsc{Billing Codes}. However, the performance using very long contexts is not substantially higher than that of using fewer tokens. Overall, performance is relatively flat across models and data selection approaches and does not exceed an F1 score of 41.07 against \textsc{Filtered diagnosis lists, or 51.03 against \textsc{Billing codes}}. For the \textsc{Filtered} list that serves as our primary target, scores across all context selection methods for GPT-5.4-mini, Mistral Medium 3, and DeepSeek V3.1 all fell within the small ranges of 2.50, 4.04, and 4.37, respectively.

For both \textsc{Filtered} and \textsc{Discharge Summary} targets, performance using the most recent notes is detrimented by using very large context amounts, while performance on \textsc{Billing Codes} demonstrates additional benefit from the additional text. A complete listing of precision, recall, and F1 scores can be found in Appendix A.

\section{Discussion}
\label{discussion}
The \textit{Imaging Procedures} task, which requires relatively shallow extraction of information data from the clinical notes, demonstrated the clearest benefit from RAG, and the performance gains from retrieval were both substantial and consistent across models.

The \textit{Antibiotic Timelines} task introduces greater complexity, requiring both temporal reasoning and clinical understanding to distinguish therapeutic antibiotics from incidental medications. While RAG also provided a significant improvement over using only recent notes, performance gains plateaued quickly—suggesting that only a limited number of passages are needed to reconstruct the key temporal history when performing targeted retrieval.

The \textit{Diagnosis Generation} task presented the greatest challenge, but we note that it is also the most subjective task. Physicians can vary in documentation practices and what is chosen to be included in the discharge summary -- an inherent limitation in automatic evaluation of this task. Performance varied by the evaluation target, with the \textsc{Billing Codes} list benefiting the most from additional text, likely due to this being a more exhaustive list of diagnoses to capture everything that can be billed. The fact that all scores, regardless of retrieval method and consistently across the models, fell within low, narrow ranges points towards performance reaching a ceiling caused by limitations of the task or evaluation method. The evaluation of this task depends on normalizing free text to ICD codes, which we do through a trained model identifying SNOMED codes before using manually written mappings. While we conducted a manual evaluation of this normalization pipeline (Appendix E) and found that core diagnoses were captured with at most minor errors in the majority of cases, this process inevitably introduces some noise into the evaluation.

We acknowledge several limitations to our study: Our evaluation draws on clinical notes from a single academic health system and the generalizability of our findings to other settings is not established by this work. Additionally, our efficiency comparison is reported in prompt tokens and does not capture the full cost of a deployed system: RAG incurs one-time chunking and embedding costs per patient as well as per-query retrieval overhead, while long-context inference can amortize repeated queries over the same record through prompt caching; end-to-end evaluation of these deployment costs remains for future work.

Furthermore, our evaluated RAG implementation uses a fixed chunking strategy, query formulation, and only two embedding models. Retrieval performance is sensitive to these parameters, and alternative configurations, or more advanced techniques such as hybrid retrieval and reranking, may yield different results. However, across all tasks and models, we observed a consistent trend: a very basic retrieval-augmented generation approach was able to closely match the performance of long-context inputs with far fewer tokens.

\section{Conclusion}
\label{conclusion}
In this work, we evaluated the effectiveness of retrieval-augmented generation across varying information demands in electronic health records across three clinical tasks. Each task was selected for its clinical relevance, reproducibility across health systems, and varying degrees of reasoning complexity.

Our results demonstrate that a simple targeted retrieval approach can reach near parity with using up to 120K tokens of recent clinical notes on these three tasks, while requiring significantly fewer input tokens. These findings show RAG's continued value even as LLMs grow more capable of processing long sequences. Further tuning the retrieval approach (queries, embedding model, retrieving more than 60 chunks, etc.), may close the remaining gap.

Future work should explore additional tasks that can be devised without extensive manual effort and informed by clinical workflows and documentation practices, as well as more granular error analyses distinguishing retrieval failures from generation failures, in order to provide a more robust assessment of models and retrieval methods over longitudinal EHR tasks.

\section*{Data Availability Statement}
The patient data underlying this article cannot be shared publicly due to ethical and legal restrictions. Intermediary analysis data will be shared upon reasonable request to the corresponding author.

\section*{Funding acknowledgement}
This work was supported by the National Library of Medicine of the National Institutes of Health under award number R01LM012973 and by the National Institute of General Medical Sciences under award number R35GM145330. The content is solely the responsibility of the authors and does not necessarily represent the official views of the National Institutes of Health.

\section*{CRediT authorship contribution statement}
\textbf{S.M.}:  Writing-original draft, Methodology, Data curation, Investigation, Conceptualization, Formal analysis, Software. \textbf{D.D.}:  Conceptualization, Methodology, Writing–review \& editing, Supervision. \textbf{T.A.M.}:  Writing–review \& editing, Methodology, Conceptualization, Funding acquisition, Supervision. \textbf{S.B.}: Formal analysis, Writing–review \& editing. \textbf{J.L.: Formal analysis, Writing–review \& editing.} \textbf{Y.G.}:  Methodology, Writing–review \& editing. \textbf{M.C.}:  Methodology, Writing–review \& editing. \textbf{A.M.}:  Methodology, Writing–review \& editing.  \textbf{M.A.}:  Supervision, Methodology, Formal analysis, Writing–review \& editing, Investigation, Conceptualization, Funding acquisition.

\section*{Declaration of competing interest}

The authors declare that they have no known competing financial interests or personal relationships that could have appeared to influence the work reported in this paper.

\section*{Ethics approval}
This study was reviewed and approved by the University of Wisconsin–Madison Institutional Review Board, with a waiver of informed consent given the retrospective use of clinical data. All data handling complied with the Health Insurance Portability and Accountability Act.

\bibliographystyle{unsrt}
\bibliography{main}

\pagebreak

\section{Appendices}
\subsection{A. Additional Results}
\setcounter{table}{0}
\renewcommand{\thetable}{A\arabic{table}}

\begin{table*}[hp!]
\centering
\resizebox{\linewidth}{!}{%
\begin{tblr}{
  cell{1}{4} = {c=3}{},
  cell{1}{7} = {c=3}{},
  cell{1}{10} = {c=3}{},
  cell{3}{1} = {r=11}{},
  cell{3}{2} = {r=3}{},
  cell{6}{2} = {r=3}{},
  cell{9}{2} = {r=5}{},
  cell{14}{1} = {r=11}{},
  cell{14}{2} = {r=3}{},
  cell{17}{2} = {r=3}{},
  cell{20}{2} = {r=5}{},
  cell{25}{1} = {r=11}{},
  cell{25}{2} = {r=3}{},
  cell{28}{2} = {r=3}{},
  cell{31}{2} = {r=5}{},
  hline{1,36} = {-}{0.08em},
  hline{2} = {3-12}{},
  hline{3,14,25} = {-}{},
  hline{6,9,17,20,28,31} = {2-12}{},
}
                 &              &                  & \textsc{Modality+Time+Location} &       &       & \textsc{Category+Time} &       &       & \textsc{Category+Time (±1 day)} &       &       \\
                 &              & \# chunks/tokens & P                      & R     & F     & P             & R     & F     & P                       & R     & F     \\
GPT-5.4-mini     & RAG (BGE)    & 20 chunks        & 48.93                  & 29.34 & 36.68 & 65.99         & 39.57 & 49.47 & 80.47                   & 48.25 & 60.33 \\
                 &              & 40 chunks        & 48.79                  & 36.54 & 41.79 & 65.41         & 48.99 & 56.02 & 77.36                   & 57.94 & 66.26 \\
                 &              & 60 chunks        & 47.04                  & 40.11 & 43.3  & 62.83         & 53.57 & 57.83 & 73.09                   & 62.31 & 67.27 \\
                 & RAG (Qwen3)  & 20 chunks        & 49.74                  & 31.97 & 38.92 & 62.41         & 40.11 & 48.83 & 72.98                   & 46.9  & 57.11 \\
                 &              & 40 chunks        & 55.47                  & 40.24 & 46.65 & 70.78         & 51.35 & 59.52 & 82                      & 59.49 & 68.95 \\
                 &              & 60 chunks        & 55.35                  & 43.14 & 48.49 & 71.24         & 55.52 & 62.41 & 81.09                   & 63.19 & 71.03 \\
                 & Recent Notes & 2560             & 36.49                  & 3.63  & 6.61  & 58.11         & 5.79  & 10.53 & 68.24                   & 6.8   & 12.36 \\
                 &              & 5120             & 40.19                  & 5.79  & 10.12 & 60.75         & 8.75  & 15.29 & 71.96                   & 10.36 & 18.12 \\
                 &              & 7680             & 47.01                  & 8.48  & 14.37 & 63.06         & 11.37 & 19.27 & 73.88                   & 13.32 & 22.58 \\
                 &              & 64K              & 52.59                  & 36.88 & 43.35 & 71.79         & 50.34 & 59.18 & 80.61                   & 56.53 & 66.46 \\
                 &              & 120K             & 50.86                  & 46.03 & 48.32 & 69.52         & 62.92 & 66.05 & 79.78                   & 72.21 & 75.8  \\
Mistral Medium 3 & RAG (BGE)    & 20 chunks        & 51.96                  & 35.67 & 42.3  & 65.59         & 45.02 & 53.39 & 78.14                   & 53.63 & 63.61 \\
                 &              & 40 chunks        & 52.52                  & 43.41 & 47.53 & 67.43         & 55.72 & 61.02 & 77.36                   & 63.93 & 70.01 \\
                 &              & 60 chunks        & 51.93                  & 48.79 & 50.31 & 65.97         & 61.98 & 63.91 & 74.93                   & 70.39 & 72.59 \\
                 & RAG (Qwen3)  & 20 chunks        & 52.94                  & 37.55 & 43.94 & 64.33         & 45.63 & 53.39 & 73.34                   & 52.02 & 60.87 \\
                 &              & 40 chunks        & 58.64                  & 45.69 & 51.36 & 72.45         & 56.46 & 63.46 & 81.52                   & 63.53 & 71.41 \\
                 &              & 60 chunks        & 58.21                  & 51.55 & 54.68 & 72.19         & 63.93 & 67.81 & 81.31                   & 72.01 & 76.37 \\
                 & Recent Notes & 2560             & 35.96                  & 4.31  & 7.69  & 51.12         & 6.12  & 10.94 & 58.99                   & 7.07  & 12.62 \\
                 &              & 5120             & 37.45                  & 6.93  & 11.7  & 53.82         & 9.96  & 16.81 & 65.82                   & 12.18 & 20.56 \\
                 &              & 7680             & 39.27                  & 10.09 & 16.06 & 50.79         & 13.06 & 20.77 & 60.73                   & 15.61 & 24.84 \\
                 &              & 64K              & 43.08                  & 40.44 & 41.72 & 56.99         & 53.5  & 55.19 & 63.87                   & 59.96 & 61.85 \\
                 &              & 120K             & 45.94                  & 43.81 & 44.85 & 60.83         & 58.01 & 59.39 & 68.88                   & 65.68 & 67.24 \\
DeepSeek V3.1    & RAG (BGE)    & 20 chunks        & 51.35                  & 33.24 & 40.36 & 66.42         & 43    & 52.21 & 81.08                   & 52.49 & 63.73 \\
                 &              & 40 chunks        & 55.29                  & 40.78 & 46.94 & 71.26         & 52.56 & 60.5  & 83.49                   & 61.57 & 70.88 \\
                 &              & 60 chunks        & 57.25                  & 45.69 & 50.82 & 72.68         & 58.01 & 64.52 & 83.05                   & 66.29 & 73.73 \\
                 & RAG (Qwen3)  & 20 chunks        & 55.22                  & 34.19 & 42.23 & 69.57         & 43.07 & 53.2  & 80                      & 49.53 & 61.18 \\
                 &              & 40 chunks        & 58.2                   & 43.94 & 50.08 & 72.64         & 54.85 & 62.5  & 82.44                   & 62.25 & 70.94 \\
                 &              & 60 chunks        & 57.82                  & 46.5  & 51.55 & 73.31         & 58.95 & 65.35 & 82.85                   & 66.62 & 73.85 \\
                 & Recent Notes & 2560             & 27.19                  & 3.97  & 6.93  & 37.33         & 5.45  & 9.51  & 48.39                   & 7.07  & 12.33 \\
                 &              & 5120             & 30.99                  & 5.92  & 9.94  & 46.13         & 8.82  & 14.8  & 55.63                   & 10.63 & 17.85 \\
                 &              & 7680             & 35.76                  & 7.94  & 13    & 51.52         & 11.44 & 18.72 & 60.91                   & 13.53 & 22.14 \\
                 &              & 64K              & 53.15                  & 35.8  & 42.78 & 74.33         & 50.07 & 59.83 & 84.42                   & 56.86 & 67.95 \\
                 &              & 120K             & 53.76                  & 45.69 & 49.4  & 73.63         & 62.58 & 67.66 & 85.19                   & 72.41 & 78.28 
\end{tblr}
}
\caption{Scores for the \textit{Imaging Procedures} task, across different models and differing amounts of provided clinical notes.}
\label{tab:imaging_additional}
\end{table*}

\begin{table*}[!htb]
\centering
\resizebox{\linewidth}{!}{%
\begin{tblr}{
  cell{1}{5} = {c=3}{},
  cell{3}{1} = {r=11}{},
  cell{3}{2} = {r=3}{},
  cell{6}{2} = {r=3}{},
  cell{9}{2} = {r=5}{},
  cell{14}{1} = {r=11}{},
  cell{14}{2} = {r=3}{},
  cell{17}{2} = {r=3}{},
  cell{20}{2} = {r=5}{},
  cell{25}{1} = {r=11}{},
  cell{25}{2} = {r=3}{},
  cell{28}{2} = {r=3}{},
  cell{31}{2} = {r=5}{},
  cell{36}{1} = {c=3}{},
  hline{1,37} = {-}{0.08em},
  hline{2} = {3-7}{},
  hline{3,14,25,36} = {-}{},
  hline{6,9,17,20,28,31} = {2-7}{},
}
                        &              &                  & \textsc{Timespan Overlap} & \textsc{Medications (name only)} &        &        \\
                        &              & \# chunks/tokens & Jaccard index    & P                       & R      & F1     \\
GPT-5.4-mini            & RAG (BGE)    & 20 chunks        & 0.4695           & 0.7232                  & 0.8491 & 0.7811 \\
                        &              & 40 chunks        & 0.4908           & 0.7139                  & 0.8849 & 0.7903 \\
                        &              & 60 chunks        & 0.4957           & 0.7072                  & 0.8865 & 0.7867 \\
                        & RAG (Qwen3)  & 20 chunks        & 0.4426           & 0.6779                  & 0.8118 & 0.7389 \\
                        &              & 40 chunks        & 0.4736           & 0.6881                  & 0.8647 & 0.7664 \\
                        &              & 60 chunks        & 0.4982           & 0.6958                  & 0.8787 & 0.7766 \\
                        & Recent Notes & 2560             & 0.3142           & 0.7176                  & 0.4821 & 0.5767 \\
                        &              & 5120             & 0.3976           & 0.751                   & 0.5816 & 0.6556 \\
                        &              & 7680             & 0.4352           & 0.7594                  & 0.6625 & 0.7076 \\
                        &              & 64K              & 0.5386           & 0.7158                  & 0.8538 & 0.7787 \\
                        &              & 120K             & 0.5308           & 0.695                   & 0.8647 & 0.7706 \\
Mistral Medium 3        & RAG (BGE)    & 20 chunks        & 0.5049           & 0.7689                  & 0.8072 & 0.7876 \\
                        &              & 40 chunks        & 0.562            & 0.7889                  & 0.8367 & 0.8121 \\
                        &              & 60 chunks        & 0.5729           & 0.7718                  & 0.8523 & 0.8101 \\
                        & RAG (Qwen3)  & 20 chunks        & 0.5004           & 0.744                   & 0.7729 & 0.7582 \\
                        &              & 40 chunks        & 0.5457           & 0.7622                  & 0.8274 & 0.7934 \\
                        &              & 60 chunks        & 0.5598           & 0.7507                  & 0.8476 & 0.7962 \\
                        & Recent Notes & 2560             & 0.3748           & 0.7738                  & 0.4044 & 0.5312 \\
                        &              & 5120             & 0.4042           & 0.774                   & 0.5272 & 0.6272 \\
                        &              & 7680             & 0.4703           & 0.7626                  & 0.6345 & 0.6927 \\
                        &              & 64K              & 0.5573           & 0.7688                  & 0.8118 & 0.7897 \\
                        &              & 120K             & 0.5405           & 0.766                   & 0.7434 & 0.7545 \\
DeepSeek V3.1           & RAG (BGE)    & 20 chunks        & 0.5441           & 0.7858                  & 0.7932 & 0.7895 \\
                        &              & 40 chunks        & 0.599            & 0.8003                  & 0.8227 & 0.8113 \\
                        &              & 60 chunks        & 0.6028           & 0.7988                  & 0.8087 & 0.8037 \\
                        & RAG (Qwen3)  & 20 chunks        & 0.5223           & 0.7557                  & 0.7745 & 0.765  \\
                        &              & 40 chunks        & 0.553            & 0.7746                  & 0.8072 & 0.7906 \\
                        &              & 60 chunks        & 0.5818           & 0.7828                  & 0.8072 & 0.7948 \\
                        & Recent Notes & 2560             & 0.3713           & 0.718                   & 0.4712 & 0.569  \\
                        &              & 5120             & 0.4614           & 0.7566                  & 0.5365 & 0.6278 \\
                        &              & 7680             & 0.5135           & 0.8012                  & 0.6081 & 0.6914 \\
                        &              & 64K              & 0.6178           & 0.7981                  & 0.7869 & 0.7925 \\
                        &              & 120K             & 0.6162           & 0.7906                  & 0.8103 & 0.8003 \\
Rule-based MAR baseline &              &                  & 0.4648           &                         &        &        
\end{tblr}
}
\caption{Scores for the \textit{Antibiotic Timelines} task, across different models and differing amounts of provided clinical notes.}
\label{tab:antibiotics_additional}
\end{table*}

\begin{table*}[!htb]
\centering
\resizebox{\linewidth}{!}{%
\begin{tblr}{
  cell{1}{4} = {c=3}{},
  cell{1}{7} = {c=3}{},
  cell{1}{10} = {c=3}{},
  cell{3}{1} = {r=11}{},
  cell{3}{2} = {r=3}{},
  cell{6}{2} = {r=3}{},
  cell{9}{2} = {r=5}{},
  cell{14}{1} = {r=11}{},
  cell{14}{2} = {r=3}{},
  cell{17}{2} = {r=3}{},
  cell{20}{2} = {r=5}{},
  cell{25}{1} = {r=11}{},
  cell{25}{2} = {r=3}{},
  cell{28}{2} = {r=3}{},
  cell{31}{2} = {r=5}{},
  hline{1,36} = {-}{0.08em},
  hline{2} = {3-12}{},
  hline{3,14,25} = {-}{},
  hline{6,9,17,20,28,31} = {2-12}{},
}
                 &              &                  & \textsc{Billing Codes} &        &        & \textsc{Discharge Summary} &        &        & \textsc{Filtered} &        &        \\
                 &              & \# chunks/tokens & P             & R      & F1     & P                 & R      & F1     & P        & R      & F1     \\
GPT-5.4-mini     & RAG (BGE)    & 20 chunks        & 0.562         & 0.3369 & 0.4213 & 0.2815            & 0.5944 & 0.382  & 0.3024   & 0.5325 & 0.3857 \\
                 &              & 40 chunks        & 0.5641        & 0.3777 & 0.4524 & 0.2743            & 0.6469 & 0.3852 & 0.2934   & 0.5771 & 0.389  \\
                 &              & 60 chunks        & 0.5706        & 0.3846 & 0.4595 & 0.2823            & 0.6704 & 0.3973 & 0.3017   & 0.5975 & 0.4009 \\
                 & RAG (Qwen3)  & 20 chunks        & 0.5955        & 0.359  & 0.448  & 0.2843            & 0.6038 & 0.3865 & 0.3024   & 0.5356 & 0.3865 \\
                 &              & 40 chunks        & 0.6029        & 0.3865 & 0.471  & 0.2848            & 0.6432 & 0.3948 & 0.2965   & 0.5583 & 0.3873 \\
                 &              & 60 chunks        & 0.5903        & 0.4009 & 0.4775 & 0.281             & 0.6723 & 0.3963 & 0.3006   & 0.5998 & 0.4005 \\
                 & Recent Notes & 2560             & 0.5909        & 0.3068 & 0.4039 & 0.3178            & 0.5812 & 0.4109 & 0.3398   & 0.5184 & 0.4105 \\
                 &              & 5120             & 0.5993        & 0.3443 & 0.4374 & 0.3135            & 0.6347 & 0.4197 & 0.327    & 0.5521 & 0.4107 \\
                 &              & 7680             & 0.5945        & 0.3705 & 0.4565 & 0.299             & 0.6563 & 0.4108 & 0.3165   & 0.5795 & 0.4094 \\
                 &              & 64K              & 0.5863        & 0.4456 & 0.5064 & 0.2647            & 0.7089 & 0.3855 & 0.2812   & 0.628  & 0.3885 \\
                 &              & 120K             & 0.5959        & 0.4462 & 0.5103 & 0.2702            & 0.7127 & 0.3918 & 0.2891   & 0.6359 & 0.3975 \\
Mistral Medium 3 & RAG (BGE)    & 20 chunks        & 0.5651        & 0.3401 & 0.4246 & 0.2657            & 0.5634 & 0.3611 & 0.2826   & 0.4996 & 0.361  \\
                 &              & 40 chunks        & 0.5692        & 0.3979 & 0.4684 & 0.26              & 0.6404 & 0.3698 & 0.2737   & 0.5623 & 0.3682 \\
                 &              & 60 chunks        & 0.5848        & 0.4294 & 0.4952 & 0.2603            & 0.6732 & 0.3754 & 0.2733   & 0.5897 & 0.3735 \\
                 & RAG (Qwen3)  & 20 chunks        & 0.5627        & 0.3638 & 0.4419 & 0.2547            & 0.5803 & 0.3541 & 0.2659   & 0.5051 & 0.3484 \\
                 &              & 40 chunks        & 0.5716        & 0.403  & 0.4727 & 0.2601            & 0.646  & 0.3709 & 0.2662   & 0.5513 & 0.359  \\
                 &              & 60 chunks        & 0.5681        & 0.4334 & 0.4917 & 0.2481            & 0.6667 & 0.3616 & 0.2579   & 0.5779 & 0.3566 \\
                 & Recent Notes & 2560             & 0.5854        & 0.3014 & 0.398  & 0.3106            & 0.5634 & 0.4004 & 0.3194   & 0.4832 & 0.3845 \\
                 &              & 5120             & 0.5954        & 0.3436 & 0.4357 & 0.3127            & 0.6357 & 0.4192 & 0.3085   & 0.5231 & 0.3881 \\
                 &              & 7680             & 0.582         & 0.383  & 0.462  & 0.292             & 0.677  & 0.408  & 0.2916   & 0.5638 & 0.3844 \\
                 &              & 64K              & 0.5011        & 0.4949 & 0.498  & 0.2121            & 0.738  & 0.3295 & 0.2159   & 0.6265 & 0.3211 \\
                 &              & 120K             & 0.5039        & 0.4667 & 0.4846 & 0.2109            & 0.6883 & 0.3229 & 0.2181   & 0.5936 & 0.319  \\
DeepSeek V3.1    & RAG (BGE)    & 20 chunks        & 0.5719        & 0.2937 & 0.3881 & 0.2891            & 0.523  & 0.3723 & 0.3046   & 0.4597 & 0.3664 \\
                 &              & 40 chunks        & 0.5868        & 0.3361 & 0.4274 & 0.2894            & 0.584  & 0.3871 & 0.3057   & 0.5145 & 0.3835 \\
                 &              & 60 chunks        & 0.6033        & 0.3643 & 0.4543 & 0.2882            & 0.6131 & 0.3921 & 0.3045   & 0.5403 & 0.3895 \\
                 & RAG (Qwen3)  & 20 chunks        & 0.5964        & 0.3017 & 0.4007 & 0.295             & 0.5258 & 0.378  & 0.3098   & 0.4605 & 0.3704 \\
                 &              & 40 chunks        & 0.6004        & 0.3555 & 0.4466 & 0.2849            & 0.5944 & 0.3852 & 0.2984   & 0.5192 & 0.379  \\
                 &              & 60 chunks        & 0.597         & 0.3739 & 0.4598 & 0.2851            & 0.6291 & 0.3924 & 0.2957   & 0.5442 & 0.3832 \\
                 & Recent Notes & 2560             & 0.5683        & 0.2639 & 0.3604 & 0.3197            & 0.523  & 0.3969 & 0.3266   & 0.4456 & 0.3769 \\
                 &              & 5120             & 0.5981        & 0.3161 & 0.4136 & 0.3202            & 0.5962 & 0.4167 & 0.3273   & 0.5082 & 0.3982 \\
                 &              & 7680             & 0.6187        & 0.3404 & 0.4391 & 0.3217            & 0.6235 & 0.4244 & 0.3319   & 0.5364 & 0.4101 \\
                 &              & 64K              & 0.5929        & 0.3955 & 0.4745 & 0.2805            & 0.6592 & 0.3935 & 0.2956   & 0.5795 & 0.3915 \\
                 &              & 120K             & 0.5802        & 0.3934 & 0.4689 & 0.2748            & 0.6563 & 0.3874 & 0.2925   & 0.5826 & 0.3894 
\end{tblr}
}
\caption{Scores for the \textit{Diagnosis Generation} task, across different models and differing amounts of provided clinical notes.}
\label{tab:diag_additional}
\end{table*}

\clearpage

\subsection{B. Prompts}
\label{sec:prompts}
\subsubsection{Imaging Procedures}
\begin{lstlisting}
# Task: Identification of Imaging Procedures from Electronic Health Records

You are an AI assistant tasked with identifying imaging procedures performed on a patient during their hospital stay. You will be provided with relevant passages from the patients Electronic Health Record.

## Instructions
1. Carefully read through all provided EHR passages and identify the imaging procedures performed and the location imaged.
2. Create a bulleted list of the imaging procedures performed during this current hospitalization, DO NOT include procedures from their history that occurred prior to this stay and DO NOT include imaging that was only performed for guidance during another procedure.
3. ONLY include procedures from these primary categories: MRI (Magnetic Resonance Imaging), CT (Computed Tomography), Ultrasound (US), X-ray (Radiograph), NM Imaging (Nuclear Medicine)
4. For each procedure found, you must identify:
- The primary imaging modality (from the list above)
- Whether there is a more specific subtype of the modality (e.g., Fluoroscopy, Mammography, Echocardiography, PET, Angiography)
- The time of the imaging (as MM/DD format or "unknown" if this cannot be determined)
- The specific body location that was imaged (e.g., brain, chest, left ankle)
5. Ignore all other medical information such as tests, medications, treatment plans, assessments, other non-imaging procedures.

## Output Format
- Use EXACTLY this format for each item: "- (MM/DD or "unknown") [Primary Imaging Modality] - [Subtype or "None"]: [Body Location]"
- Use "None" in place of subtype when no specific subtype is necessary and "unknown" in place of the '
date if the time of the imaging cannot be determined.
- Present as a clean bulleted list
- Include no explanatory text, introductions, or conclusions
- Do not number the items
- If multiple imaging procedures of the same type were performed on different locations, list each separately
- The same imaging occurence may be mentioned multiple times throughout the EHR, only include one entry per occurence.
- If no imaging procedures exist, output only: "No imaging procedures identified."

## Example
2018-03-12 15:43:00 H&P
Patient admitted with chest pain. Cardiac enzymes were elevated. A chest X-ray was performed on 3/10 showing cardiomegaly. CT scan of the chest was completed to rule out aortic dissection. The patient also underwent a transthoracic echocardiogram today.

Example output:
- (03/10) X-ray - None: Chest
- (unknown) CT - None: Chest
- (03/12) Ultrasound - Echocardiogram: Heart

## Begin Task
EHR passages:
[INSERT TEXT]

Your response as a list of imaging types and locations:
\end{lstlisting}
\subsubsection{Antibiotic Timelines}
\begin{lstlisting}
# Task: Identification of Administered Antibiotics and Date Ranges from Electronic Health Records\n\n
You are an AI assistant tasked with identifying antibiotics administered to a patient during
their hospital stay and the date ranges for each antibiotic's use. You will be
provided with relevant passages from the patient's Electronic Health Record (EHR), each with an
associated timestamp.\n\n
## Instructions\n
1. Carefully read through all provided EHR passages, noting their timestamps.\n
2. Create a list of antibiotics being administered, prescribed, or continued.\n
3. Do not include antibiotics given for prophylaxis or minor conditions. Only include antibiotics being used for the treatment of the major acute condition of the ICU patient.\n
4. For each antibiotic, determine the start and end dates of its use by inferring from the timestamps of the passages and any date information within the text.\n
5. Use the format MM/DD-MM/DD for date ranges. If the antibiotic use is ongoing, use \"present\" for the end date.\n
6. Don't include dosages or administration routes.\n
7. If a date range can't be determined whatsoever, list the antibiotic with \"(dates unclear)\" after it.\n
\n
## Output Format\n
Provide your response as a list of antibiotics with their date ranges in the following format:\n
- Antibiotic 1 (MM/DD - MM/DD)\n
- Antibiotic 2 (MM/DD - present)\n
## Example\n
Right now it is 2019-09-15 14:51:00.\n
EHR passages:\n\n
2019-09-12 10:15:00\n
\"Patient admitted with suspected pneumonia. Started on IV ceftriaxone 1g daily.\"\n\n
2019-09-14 14:30:00\n
\"Blood cultures positive for MRSA. Ceftriaxone discontinued. Started on IV vancomycin 1g q12h.\"\n\n
Output:\n
- Ceftriaxone (09/12-09/14)\n
- Vancomycin (09/14-ongoing)\n\n
## Begin Task\n
Right now it is [TIMESTAMP].\n\nEHR passages:\n\n[INSERT TEXT]
\n\nYour response as a list of antibiotic names and date ranges:\n
\end{lstlisting}
\subsubsection{Diagnosis Generation}
\begin{lstlisting}
# Task: Identification of Clinically Important Diagnoses
You are an AI assistant tasked with creating a clinically relevant problem list for an ICU patient's stay. You will analyze passages of clinical notes from their hospitalization and identify diagnoses  that required active management or monitoring during their stay.

# Task
Review the provided clinical note passages and generate a structured list of diagnoses that:
1. Required active management during the hospitalization
2. Were clinically significant to their critical care course
3. Impacted their ICU care plan or outcomes

# Inclusion Criteria
Include diagnoses that meet ANY of these criteria:
- Acute conditions requiring ICU-level care (e.g., NSTEMI, septic shock, acute respiratory failure)
- Chronic conditions requiring active management or affecting ICU care (e.g., atrial fibrillation, COPD exacerbation)
- New diagnoses made during the encounter
- Complications that developed during the stay
- Conditions requiring monitoring or intervention (e.g., acute kidney injury, severe electrolyte disorders)
- Neurologic/cognitive conditions affecting ICU care (e.g., delirium, acute stroke)
- Conditions directly related to the reason for ICU admission

# Exclusion Criteria
Do NOT include:
- Stable chronic conditions not requiring active management (e.g., well-controlled diabetes, stable hypothyroidism)
- Historical diagnoses not affecting current care (e.g., "history of appendectomy")
- Social history items (e.g., "former smoker")
- Procedural or post-surgical statuses (e.g., "s/p CABG", "post-cholecystectomy")
- Symptoms without clear diagnoses
- Conditions that resolved prior to admission
- Incidental findings not requiring intervention

# Output Format
Present the diagnoses as a numbered list, ordered by clinical priority :
1. [Primary diagnosis]
2. [Secondary diagnosis]
3. [Tertiary diagnosis]
...
# Note
- Every listed item must be a specific medical diagnosis - Use standard medical terminology for diagnoses

# EHR passages:\n[INSERT TEXT]\n
# Output the diagnoses that required active management or monitoring during their ICU stay, as instructed.  Every listed item must be a specific medical diagnosis:
\end{lstlisting}

\subsection{C. SNOBERT Training}
\label{sec:snobert}

We trained SNOBERT using the configuration provided in the authors' Github Repository as-is with the same training data from the SNOMED CT Entity Linking Challenge, but using the International SNOMED vocabulary files from 2025, since we did not have access to the version used for the challenge. We trained a single model, as opposed to their approach of ensembling six with varying class weights and data splits for the competition, but through expert review, we determined performance on the downstream ICD-10 code extraction step to be acceptable. 

\subsection{D. \textsc{Filtered} Diagnoses Validation}

To assess the reliability of the Gemma-3-32B filtering step, in which the full ICD-10 billing code list is reduced to the diagnoses directly relevant to the text in the discharge summary, a medical physcian independently performed the same filtering task on a random sample of 20 hospitalizations. Given the discharge summary text and the unfiltered list of billing codes, the physcian produced their own filtered list under the same instructions as Gemma-3-32B. The model's output was then evaluated against this manual annotation, resulting in 86.72 precision, 92.50 recall, and 89.52 F1. We interpret this as evidence that the \textsc{Filtered} gold standard reasonably approximates clinician judgment for this filtering task.

\subsection{E. Free-text to ICD-10 Normalization}

To assess the reliability of the text normalization pipeline used in the \textit{Diagnosis Generation} task (comprising SNOBERT-based SNOMED extraction followed by the SNOMED-to-ICD-10 mapping from the SNOMED CT release), a medical physcian manually reviewed the pipeline's output on 40 cases drawn from our test set: 20 discharge summary normalizations and 20 LLM-generated output normalizations. For each case, the physcian was provided with the source free text and the resulting ICD-10 code list, and assigned a score from 1 to 5 reflecting how faithfully the normalized codes captured the clinically significant diagnoses present in the text. The rubric descriptions are shown in Table \ref{tab:rubric} and the distribution of scores is shown in Table \ref{tab:normalization_eval}.

\begin{table}[h]
\centering
\begin{tabular}{cp{0.75\textwidth}}
\toprule
Score & Description \\
\hline
5 & Essentially correct. All clinically significant codes captured; any errors limited to minor specificity differences (e.g., laterality, episode of care) that would not change clinical interpretation or reimbursement. \\
4 & Minor errors. Core diagnoses captured correctly, with one or two minor omissions or minor specificity mismatches. \\
3 & Partially correct. Most major diagnoses present, but with at least one clinically meaningful omission or a code substitution within the same clinical domain (e.g., Type 1 vs.\ Type 2 diabetes). \\
2 & Substantially incomplete or inaccurate. Multiple major omissions, or one or more codes in the wrong clinical domain entirely. \\
1 & Largely wrong. The extracted list bears little resemblance to the reference set --- most major diagnoses missing, fabricated codes, or systematic misclassification. \\
\bottomrule
\end{tabular}
\caption{Rubric for manual evaluation of free-text to ICD-10 normalization.}
\label{tab:rubric}
\end{table}

\begin{table}[h]
\centering
\resizebox{0.4\linewidth}{!}{%
\begin{tblr}{
  column{2} = {c},
  column{3} = {c},
  hline{1,8} = {-}{0.08em},
  hline{2,7} = {-}{},
}
Score & Discharge Summaries & LLM Outputs \\
5     & 5 (25\%)            & 9 (45\%)    \\
4     & 6 (30\%)            & 5 (25\%)    \\
3     & 6 (30\%)            & 5 (25\%)    \\
2     & 3 (15\%)            & 1 (5\%)     \\
1     & 0 (0\%)             & 0 (0\%)     \\
Mean  & 3.65                & 4.10        
\end{tblr}
}
\caption{Manual evaluation of the SNOBERT + SNOMED-to-ICD-10 normalization pipeline. Scores of 4 or higher indicate that core diagnoses were captured with at most minor errors.}
\label{tab:normalization_eval}
\end{table}

Cases with a score of 4 or higher (indicating that core diagnoses were captured correctly with at most minor specificity issues) accounted for 55\% of discharge summary normalizations and 70\% of LLM output normalizations. No cases were rated as largely wrong (score 1). Notably, normalization quality was somewhat higher on LLM outputs than on discharge summaries, likely because LLM outputs tend to use more explicit, standardized diagnostic phrasing, whereas discharge summaries contain more abbreviations, narrative phrasing, and inferred diagnoses that pose challenges for the SNOBERT model.

\end{document}